\documentclass{article}

\usepackage{microtype}
\usepackage{graphicx}
\usepackage{subfigure}
\usepackage{booktabs} 

\usepackage{hyperref}

\usepackage[accepted]{icml2025}

\usepackage{amsmath}
\usepackage{amssymb}
\usepackage{mathtools}
\usepackage{amsthm}

\usepackage{multirow}
\usepackage{algorithm}
\usepackage{algorithmic}
\usepackage{enumitem}
\def\ZZ{\mathbb{Z}}
\def\RR{\mathbb{R}}
\def\NN{\mathbb{N}}

\usepackage[capitalize,noabbrev]{cleveref}

\theoremstyle{plain}

\theoremstyle{definition}

\theoremstyle{remark}

\icmltitlerunning{Batch-Max: Higher LLM Throughput}

\begin{document}

\twocolumn[
\icmltitle{Batch-Max: Higher LLM Throughput using\\Larger Batch Sizes and KV Cache Compression}




\begin{icmlauthorlist}
\icmlauthor{Michael R. Metel}{1}
\icmlauthor{Boxing Chen}{1}
\icmlauthor{Mehdi Rezagholizadeh}{2}
\end{icmlauthorlist}
\icmlaffiliation{1}{Huawei Noah's Ark Lab}
\icmlaffiliation{2}{Advanced Micro Devices, Inc.}

\icmlcorrespondingauthor{Michael R. Metel}{michael.metel@h-partners.com}

\icmlkeywords{LLM throughput,KV cache compression,batched inference,resource-constrained inference}

\vskip 0.3in
]



\printAffiliationsAndNotice{}  

\begin{abstract}
Several works have developed eviction policies to remove key-value (KV) pairs from the KV cache for more efficient inference. The focus has been on compressing the KV cache after the input prompt has been processed for faster token generation. In settings with limited GPU memory, and when the input context is longer than the generation length, we show that by also compressing the KV cache during the input processing phase, larger batch sizes can be used resulting in significantly higher throughput while still maintaining the original model's accuracy. 
\end{abstract}

\section{Introduction}

This work is focused on increasing LLM inference throughput using limited GPU memory more efficiently in scenarios where it is expected that the input context will be longer than the generation length, such as for summarization tasks or in-context learning. Attention-based LLM inference is comprised of two stages: processing the input context (prefilling), where the input's KV cache is computed and the first new token is generated, and token generation (decoding), where one token is generated per forward pass of the model. Prefilling can be computed in parallel, making it compute bound, whereas decoding is memory bandwidth bound, as it requires reloading the KV cache each generation step \cite{shazeer2019}.  

Several works (see Section \ref{lit_review}) have proposed KV cache eviction policies to remove unimportant KV pairs from the KV cache during decoding, where in general, the KV cache size $|kv|$ is restricted to a maximum size of $\overline{|kv|}$ KV pairs per attention head and batch sample. 

There are cases where only compressing the KV cache during decoding can maximize throughput. Given a fixed batch size $b$ and sufficient GPU memory to perform prefilling, KV cache compression should only be done during decoding: We want to process the entire input prompt in parallel during prefilling, and transfer the least amount of data during decoding. Another example is when the input sequence length $s<\overline{|kv|}$, which can occur when the input context is much shorter than the expected generation length.

This work concentrates on tasks where $s$ is expected to be larger than the generation length, with a fixed GPU memory budget, and the freedom to choose $b$. When $s>\overline{|kv|}$ and the KV cache is only compressed during decoding, the GPU memory used per attention head for $(s-\overline{|kv|})b$ KV pairs during prefilling will be left idle during decoding. This most importantly limits the maximum $b$ which can be used. Restricting the KV cache size to $\overline{|kv|}$ during both prefilling \& decoding (P\&D) then enables higher GPU usage and throughput by being able to increase $b$. 

Compressing the KV cache during prefilling creates new challenges and potential trade-offs:

\begin{enumerate}[itemsep=-3pt]
	\item Slower prefilling: The input prompt must now be processed in a block-wise manner while using a KV cache eviction algorithm. 	
	\item KV pair error: After the first block, error will exist in the non-evicted KV pairs of the input prompt, being computed using the compressed KV cache of past input prompt tokens.  	
	\item Suboptimal KV pair eviction: All of the input prompt KV pairs can no longer be observed before deciding which KV pairs to evict.    
\end{enumerate}

Numerical experiments, consisting of different tasks, LLM architectures and GPU models, show that the ability to increase $b$ by using P\&D KV cache eviction outweighs any decrease in speed or accuracy incurred by the challenges listed above. Significantly higher throughput ($44.0\%$ higher on average) was achieved compared to an upper bound on the throughput using decoding-only compression, while maintaining the accuracy of the full KV cache model ($2.2\%$ lower on average). 

In what follows, Section \ref{lit_review} summarizes the literature on KV cache eviction methods, Section \ref{method} describes Batch-Max (BM), a candidate P\&D KV cache eviction method, Section \ref{experiments} describes the experimental setup and results, with the paper concluding in Section \ref{conclusion}.  

\section{Literature Review}
\label{lit_review}

The papers $H_2O$ \cite{zhang2023}, SqueezeAttention \cite{wang2024}, FastGen \cite{ge2024}, SnapKV \cite{li2024}, Scissorhands \cite{liu2023}, TOVA \cite{oren2024}, and SimLayerKV \cite{zhang2024} propose different decoding-only KV cache eviction methods. 

After prefilling, $H_2O$ compresses the KV cache, then adds and removes one KV pair every generation step, keeping the KV cache split evenly between a window of the most recent KV pairs and those with the largest sum of past attention scores.

SqueezeAttention uses different KV cache sizes per layer, measuring their importance during prefilling based on the cosine similarity between their input and output. Their method was tested with different eviction rules including $H_2O$.

FastGen selects one of several different KV cache compression policies (including $H_2O$) to compress each attention head by minimizing GPU memory usage while ensuring a minimum recovery of attention weights, based on the KV cache of the input prompt. 

SnapKV only compresses the KV cache one time after prefilling using an eviction rule similar to $H_2O$, keeping a window $w$ of recent KV pairs and the KV pairs with the largest sum of attention weights over $w$, which go through a pooling layer to avoid sparse selections. 

Scissorhands also keeps a window of recent KV pairs but evicts older pairs based on how often their attention scores are below average over a window of past tokens. This work also considers using different KV cache sizes per layer, as well as only evicting KV pairs every $t>1$ generation steps. 

TOVA simply removes the token with the lowest attention score based on the current query, which benefits from not being biased (see Section \ref{ave_att} for more discussion).

SimLayerKV identifies ``lazy" layers which follow the attention pattern discovered in StreamingLLM \cite{xiao2024}. If the average attention given to recent tokens and the first four ``attention sink" tokens surpasses a threshold, the layer is deemed lazy, with only these KV pairs being kept during the decoding phase, with non-lazy layers keeping their full KV cache.   

EasyKV \cite{ren2024} proposes an eviction policy, RoCo, which splits the KV cache between pairs with the highest attention weight standard deviations and highest average attention weights. KV cache compression is considered during prefilling, decoding, or in both stages, depending on the input and generation lengths, e.g. only prefilling compression is performed for CNN/DM (see Section \ref{experiments}), since the input should be longer than the generation length for this summarization task. Compression is done by processing and evicting KV pairs in a block-wise manner when prefilling, and adding and removing one KV pair per generation step. The focus of EasyKV was on the improved accuracy of RoCo compared to other eviction rules (including $H_20$, ScissorHands, and TOVA), with no experiments using $b>1$, or any attempt to examine the effect on throughput using P\&D compression. 

\section{Batch-Max}
\label{method}

An implementation of P\&D KV cache compression is now described. Let $s_j$ equal the input sequence length of samples $j=1,...,b$, and $\overline{s}:=\max\limits_{j} s_j$. To perform inference in parallel, $\overline{s}-s_j$ pad tokens are added from the left to each sample $j$. The size of the KV cache is restricted to $\overline{|kv|}$ KV pairs per attention head and sample.  

\subsection{P\&D KV Cache Eviction}

After processing the first block of input tokens during prefilling, which can equal up to $\overline{|kv|}$ tokens, KV pairs are evicted every $p\in\NN$ tokens, where once $|kv|=\overline{|kv|}$, $p$ KV pairs are removed, see Algorithm \ref{alg1}.

\begin{algorithm}
	\caption{P\&D KV Cache Eviction} 
	\begin{algorithmic}     
		\STATE {\bfseries Input:} $\overline{s}\in\NN$: padded input sequence length of all samples; $\overline{|kv|}\in\NN$: maximum KV cache size per head and sample; $p(=64)\in\NN$: KV cache eviction amount; $\text{max\_gen}\in\NN$: maximum number of generated tokens	   
		\STATE {\bfseries Prefilling:}			
		\STATE $t_1=\min(\overline{s},\overline{|kv|})$
		\STATE process tokens $[0,t_1-1]$
		\WHILE {$t_1<\overline{s}$}
		\STATE remove $p$ KV pairs
		\STATE $t_0=t_1$
		\STATE $t_1=\min(\overline{s},t_0+p)$		
		\STATE process tokens $[t_0,t_1-1]$		 
		\ENDWHILE		
		\STATE {\bfseries Decoding:}  
		\FOR{$t=1$ {\bfseries to} max\_gen$-1$}%
		\IF{$|kv|=\overline{|kv|}$}
		\STATE remove $p$ KV pairs
		\ENDIF
		\STATE generate token $\overline{s}+t-1$
		\ENDFOR
	\end{algorithmic}
	\label{alg1}	
\end{algorithm}

\subsection{Average Attention Eviction Rule} 
\label{ave_att}

$H_20$ ranks KV pairs for eviction based on the sum of their past attention weights. Considering a zero-initialized (variable-sized) vector $sum\_weights\in \RR^{|kv|}$ for each attention head and sample, in each forward pass with a block of inputs of length $I\in\NN$, let it be updated as 
$$sum\_weights+=\sum_{i=1}^{I} attn\_weights[i],$$
where $attn\_weights[i]\in \RR^{|kv|}$ contains the attention weights for the query generated from the $i^{th}$ input in the block. For example, in Algorithm \ref{alg1}, when $t_1=t_0+p$ during prefilling, $I=p$, and during decoding $I=1$. Let $kv\_ids\in\ZZ^{|kv|}_{\geq 0}$ be the position IDs of the tokens the KV pairs were generated from, with $curr\_id\in\ZZ_{\geq 0}$ being the current position ID. During prefilling $curr\_id=t_1-1$ and during decoding $curr\_id=\overline{s}+t-1$.  

Ranking based on $sum\_weights$ is biased towards KV pairs generated from earlier tokens given that the entries of earlier KV pairs are the sum of more (i.e. $curr\_id+1-kv\_ids[i]$) $attn\_weights$ vectors. In addition, the average value in $attn\_weights$ generated from a token with position ID $i$ equals $\frac{1}{min(i+1,|kv|)}$, which is larger for earlier tokens. For example, the first computed KV pair gets an initial $sum\_weights$ value of 1 and is the sum of $curr\_id+1$ $attn\_weights$, whereas the $sum\_weights$ value of the most recent token only equals $\frac{1}{min(curr\_id+1,|kv|)}$ if it receives the average value of $attn\_weights$. 

This bias has been remedied by not evicting a window of recent KV pairs in past works. We instead evict KV pairs with the smallest average attention weights,   
$$ave\_weights = \frac{sum\_weights}{curr\_id+1-kv\_ids},$$
where the division is done element-wise. This simple eviction rule directly corrects for the mentioned bias without having to separate KV pairs based on recency. This eviction rule also forms a part of RoCo \cite{ren2024} which, in addition, protects KV pairs from eviction based on the standard deviation of their attention weights. Simply evicting based on $ave\_weights$ was found to maintain sufficient accuracy, while also making our experiments clearer by using a hyperparameter-free eviction rule.

\begin{algorithm}
	\caption{Extreme Decoding-only KV Cache Eviction} 
	\begin{algorithmic}     
		\STATE {\bfseries Input:} $\overline{s}\in\NN$: padded input sequence length of all samples; $\overline{|kv|}\in\NN$: maximum KV cache size per head and sample; $\text{max\_gen}\in\NN$: maximum number of tokens to generate	   
		\STATE {\bfseries Prefilling:}					
		\STATE process tokens $[0,\overline{s}-1]$		 		
		\STATE remove all but the most recent KV pair	
		\STATE {\bfseries Decoding:}  
		\FOR{$t=1$ {\bfseries to} max\_gen$-1$}%
		\STATE generate token $\overline{s}+t-1$
		\IF{$|kv|=\overline{|kv|}$}
		\STATE remove all but the most recent KV pair		
		\ENDIF		
		\ENDFOR
	\end{algorithmic}
	\label{alg3}	
\end{algorithm}

\section{Experiments}
\label{experiments}

Our goal is to observe if higher throughput can be achieved by using P\&D KV cache eviction compared to decoding-only eviction. Our candidate method for P\&D compression is Batch-Max (BM), using Algorithm \ref{alg1} with the average attention eviction rule described in Section \ref{ave_att}. Instead of trying every variation of the methods described in Section \ref{lit_review}, we consider a form of extreme decoding-only compression (ED, Algorithm \ref{alg3}) which gives an upper bound on the potential throughput decoding-only compression can produce. ED uses the simplest eviction rule, by only keeping the most recent KV pair, and with $\overline{|kv|}=2$, it always loads the smallest non-empty KV cache, making it the fastest possible decoding-only KV cache eviction algorithm. 

The throughput of BM is compared with ED, while keeping its accuracy close to the full KV cache model (FKV). Experiments were performed on three tasks: CNN/DM (1-shot, \citealt{nallapati2016}), NarrativeQA (2-shot, \citealt{kocisky2018}), and GSM8K (16-shot, \citealp{cobbe2021}). Two LLM architectures on different GPU models were used: Llama-2-13b-chat \cite{touvron2023} on 4 (CNN/DM \& NarrativeQA) or 2 (GSM8K) NVIDIA V100 (32GB) GPUs, and Phi-3.5-mini-instruct (3.8B, \citealp{abdin2024}) on 4 NVIDIA TITAN V (12GB) GPUs. 

\subsection{Experimental Procedure \& Analysis}
In all experiments, the smallest batch size $b^0$ such that ED with $\overline{|kv|}=2$ ran out of memory was found. Using a batch size of $b=b^0-1$, the highest possible throughput using ED was computed, as well as the accuracy of FKV. We then tried to maximize the throughput of BM by increasing $b$, while maintaining the same level of accuracy as FKV by keeping  $\overline{|kv|}$ sufficiently large. The choice of $p=64$ for BM was used for all experiments, which was found to reasonably balance speed (eviction every $p$ processed/generated tokens) and accuracy ($|kv|\geq \overline{|kv|}-p$). For the Llama-2 experiments in Table \ref{T1}, ED with $\overline{|kv|}=65$ was also tested, which corresponds to $p=64$ in BM, to ensure that the throughput with $\overline{|kv|}=2$ was higher. 

We were able to consistently generate higher throughput using BM compared to ED. In the Llama-2 experiments in Table \ref{T1}, two results for BM are given for each task, one keeping the rouge-2 or accuracy always slightly greater than FKV, where the throughput was on average $38.0\%$ higher than ED, and the other keeping the accuracy of BM near FKV, where on average the throughput is $50.4\%$ higher than ED, and the accuracy is on average $98.0\%$, and at least equal to $96.3\%$ of FKV's accuracy. In Table \ref{T2}, the experiments using Phi-3.5 are given, where on average the throughput is $37.7\%$ higher than ED, and the accuracy is on average $97.6\%$, and at least equal to $96.5\%$ of FKV's accuracy.  

\begin{table}
	\centering
	\begin{tabular}{|lcccc|}
		\cline{1-3}	
		Task: &\multicolumn{2}{c|}{CNN/DM} \\	
		\cline{4-5}				
		Method & $b$ & $\overline{|kv|}$ & rouge-2 & tokens/s \\
		\hline				
		ED & 5 & 2 & OOM & OOM \\
		ED & 4 & 2 & 0.000 & {\bf 42.0} \\		
		ED & 4 & 65 & 0.003 & 41.5 \\				
		FKV & 4 & N/A & {\bf 0.145} & 30.3 \\		
		BM & 32 & 1024 & {\bf 0.146} & {\bf 73.8} \\		
		BM & 40 & 896 & {\bf 0.142} & {\bf 80.3} \\				
		\hline				          	
		Task: &\multicolumn{2}{c|}{NarrativeQA} \\	
		\cline{4-5}			
		Method & $b$ & $\overline{|kv|}$ & rouge-2 & tokens/s \\
		\hline				
		ED & 5 & 2 & OOM & OOM \\	
		ED & 4 & 2 & 0.000 & {\bf 40.0} \\				
		ED & 4 & 65 & 0.001 & 39.3\\												
		FKV & 4 & N/A &{\bf 0.312} & 26.4 \\	
		BM & 15 & 1792 & {\bf 0.314} & {\bf 43.3}\\
		BM & 16 & 1728 & {\bf 0.310} & {\bf 46.2} \\					
		\hline				          
		Task: &\multicolumn{2}{c|}{GSM8K} \\	
		\cline{4-5}				
		Method & $b$ & $\overline{|kv|}$ & accuracy & tokens/s \\
		\hline				
		ED & 4 & 2 & OOM & OOM \\	
		ED & 3 & 2 & 0.000 & {\bf 30.3} \\	
		ED & 3 & 65 & 0.000 & 30.1 \\											
		FKV & 3 & N/A & {\bf 0.340} & 22.9 \\	
		BM & 8 & 1536 & {\bf 0.341} & {\bf 39.3} \\	
		BM & 10 & 1408 & {\bf 0.327} & {\bf 43.7} \\					
		\hline				          
	\end{tabular}        
	\caption{Llama-2-13b-chat experiments comparing ED (Alg. \ref{alg3}), full KV cache (FKV), and Batch-Max (BM). Relevant values to compare are in bold.}
	\label{T1}
	\vspace{-10pt}
\end{table}	

\begin{table}
	\centering
	\begin{tabular}{|lcccc|}
		\cline{1-3}	
		Task: &\multicolumn{2}{c|}{CNN/DM} \\	
		\cline{4-5}				
		Method & $b$ & $\overline{|kv|}$ & rouge-2 & tokens/s \\
		\hline				
		ED & 3 & 2 & OOM & OOM \\		
		ED & 2 & 2 & 0.001 & {\bf 29.0} \\	
		FKV & 2 & N/A & {\bf 0.153} & 27.0 \\	
		BM & 5 & 2176 & {\bf 0.148} & {\bf 36.3} \\				
		\hline				          	
		Task: &\multicolumn{2}{c|}{NarrativeQA} \\	
		\cline{4-5}				
		Method & $b$ & $\overline{|kv|}$ & rouge-2 & tokens/s \\
		\hline		
		ED & 3 & 2 & OOM & OOM \\			
		ED & 2 & 2 & 0.000 & {\bf 28.0} \\							
		FKV & 2 & N/A & {\bf 0.360} & 26.5 \\			
		BM & 6 & 1984 & {\bf 0.351} & {\bf 37.4} \\					
		\hline				          
		Task: &\multicolumn{2}{c|}{GSM8K} \\	
		\cline{4-5}				
		Method & $b$ & $\overline{|kv|}$ & accuracy & tokens/s \\
		\hline		
		ED & 3 & 2 & OOM & OOM \\			
		ED & 2 & 2 & 0.000 & {\bf 28.4} \\			
		FKV & 2 & N/A & {\bf 0.783} & 27.1 \\											
		BM & 8 & 1664 & {\bf 0.774} & {\bf 43.8} \\					
		\hline				          
	\end{tabular}        
	\caption{Phi-3.5-mini-instruct experiments comparing ED (Alg. \ref{alg3}), full KV cache (FKV), and Batch-Max (BM). Relevant values to compare are in bold.}
	\label{T2}
	\vspace{-10pt}
\end{table}

\subsection{Further Details}

In order to fairly measure throughput, 512 tokens were always generated, ignoring any EOS tokens. Llama-2's maximum sequence length is 4096, whereas Phi-3.5 supports up to 128K tokens. For consistency, we limited the maximum input length to $3584=4096-512$ for all experiments, which only affected the CNN/DM dataset. The evaluation was performed on 960 same-seed randomly chosen test set samples, which is divisible by $D:=\{1,2,3,4,5,6,8,10,12,15,16,20,24,30,32,40,48,...\}$. Only using batch sizes $b\in D$ ensured that all experiments were evaluated on the exact same samples. In all experiments, ED ran out of memory with $b^0\leq 5$, resulting in there being no restriction from choosing $b\in D$. For all but one experiment (Phi-3.5 CNN/DM) $\max(b)+1\notin D$ when using BM. In practice, when not trying to fairly compare with ED and FKV, higher throughput can be expected by freely maximizing $b\in\NN$. When choosing $\overline{|kv|}$ for BM, multiples of 128 $\{896, 1024, 1408, 1536, 1664, 1792, 2176\}$ were tried, which was further refined in two experiments to multiples of 64 $\{1728, 1984\}$. 

\section{Conclusion}
\label{conclusion}

With the goal of maximizing LLM inference throughput, the use of KV cache eviction during both the prefilling and decoding phases was explored. A simple implementation was proposed, Batch-Max, using an average attention eviction policy, which was able to significantly increase the throughput compared to an upper bound on the throughput using any decoding-only KV cache eviction method, while maintaining the accuracy of the full KV cache model. Our experiments indicate that in settings with limited GPU memory and where input sequences are expected to be longer than the generation length, KV cache compression during both prefilling and decoding should be used given that it enables larger batch sizes, resulting in higher throughput, while not incurring significant accuracy degradation.

\section*{Impact Statement}

This paper presents work whose goal is to advance the field of 
Machine Learning. There are many potential societal consequences 
of our work, none which we feel must be specifically highlighted here.

\bibliographystyle{icml2025}
\bibliography{batchmax}

\end{document}